\documentclass[10pt]{article}

\PassOptionsToPackage{numbers, compress}{natbib}
\usepackage{natbib}
\usepackage[utf8]{inputenc}
\usepackage[a4paper, total={6in, 10in}]{geometry}
\usepackage{graphicx}
\usepackage{indentfirst}
\usepackage{authblk}
\usepackage{lineno}
\usepackage{adjustbox}

\usepackage{graphicx}
\usepackage{subcaption}
\usepackage{comment}

\title{A framework for dynamically training and adapting deep reinforcement learning models to different, low-compute, and continuously changing radiology deployment environments}
\author[2]{Guangyao Zheng}
\author[1]{Shuhao Lai}
\author[2]{Vladimir Braverman}
\author[3,4]{Michael A. Jacobs}
\author[5]{Vishwa S. Parekh}
\affil[1]{Department of Computer Science, The Johns Hopkins University, Baltimore, MD 21208, USA}
\affil[2]{Department of Computer Science, Rice University, Houston, TX, USA}
\affil[3]{Department of Diagnostic And Interventional Imaging, McGovern Medical School, UTHealth Houston, Houston, TX, USA}
\affil[4]{Department of Radiology, Johns Hopkins University School of Medicine, Baltimore, MD 21205, USA}
\affil[5]{University of Maryland Medical Intelligent Imaging (UM2ii) and Department of Diagnostic Radiology and Nuclear Medicine, University of Maryland School of Medicine, Baltimore, MD 21201, USA}

\begin{document}

\maketitle
\begin{abstract}
While Deep Reinforcement Learning has been widely researched in medical imaging, the training and deployment of these models usually require powerful GPUs. Since imaging environments evolve rapidly and can be generated by edge devices, the algorithm is required to continually learn and adapt to changing environments, and adjust to low-compute devices. To this end, we developed three image coreset algorithms to compress and denoise medical images for selective experience replayed-based lifelong reinforcement learning. We implemented neighborhood averaging coreset, neighborhood sensitivity-based sampling coreset, and maximum entropy coreset on full-body DIXON water and DIXON fat MRI images. All three coresets produced 27x compression with excellent performance in localizing five anatomical landmarks: left knee, right trochanter, left kidney, spleen, and lung across both imaging environments. Maximum entropy coreset obtained the best performance of $11.97\pm 12.02$ average distance error, compared to the conventional lifelong learning framework's $19.24\pm 50.77$.

\end{abstract}

\section{Main}
Deep Reinforcement Learning (DRL) is a sub-branch of machine learning algorithms that has the basis of reinforcement learning incorporated with deep learning. DRL's unique ability to explore the environment and learn to perform a task efficiently is similar to human learning process and is suitable for applications to the medical imaging field through the exploration of imaging environments. DRL has achieved remarkable success in recent years in the medical imaging field, with diverse applications to disease classification, image segmentation, anatomical landmark localization \cite{ghesu2017multi,tseng2017deep,maicas2017deep,ma2017multimodal,alansary2018automatic,ali2018lung,alansary2019evaluating,vlontzos2019multiple,Allioui2022MultAgent,Zhang2021wholeprocess,Stember2022drl}. 

Although DRL has yielded excellent results in medical imaging, since DRL relies on a deep neural network to learn complex representations and patterns from the environment, the model is usually large-scale and not easy to train. Thus most researchers and medical professionals use high-end GPUs or cloud computing which has high-end GPUs to train these models efficiently. Moreover, technological advancements have allowed the production of higher-resolution images, further increasing the computational infrastructure requirement. However, to deploy these models in the real world, not all users have access to such computational power caused of limitations in funding or location, making DRL deployment in the real-world prohibitively expensive and inaccessible.

\begin{figure}[htp]
    \centering
    \includegraphics[width=15cm]{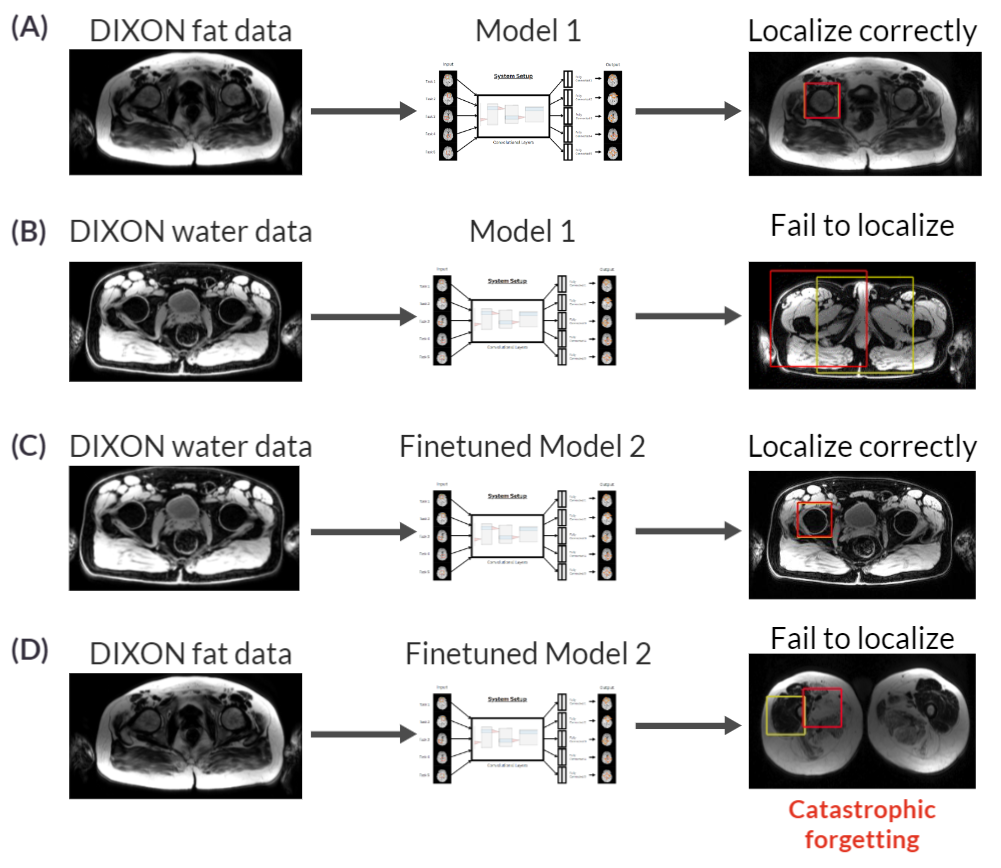}
    \caption{Illustration of catastrophic forgetting. (A) Model 1 is trained on DIXON fat data, and it is able to localize the right trochanter. (B) When Model 1 encounters DIXON water data, it cannot localize the right trochanter. (C) When we finetune Model 1 on DIXON water data to produce Model 2, it is able to localize the right trochanter on DIXON water data. (D) But the model is unable to localize the right trochanter on DIXON fat data, on which it was initially trained.}
    \label{fig:figure1}
\end{figure}

Additionally, as medical imaging technology continues to evolve, an increasing number of imaging environments will need to be explored. Currently, we have X-Ray, CT, PET, Ultrasound, and MRI. An MRI alone has many imaging sequences such as T1, T1-post contrast, Fluid attenuated inversion recovery(FLAIR), and T2. Medical images also have diverse orientations with images in the Axial, Coronal, or Sagittal views. Pathology is also diverse, such as high-grade glioma (HGG) or low-grade glioma (LGG) for brain tumors. More diverse imaging environments will emerge and training a model for each environment and each task would not only be inefficient but also impossible to manage and maintain. However, a DRL model trained and deployed on an older environment and task cannot do well on a new, unseen environment and task. And finetuning this model to the new environment and task will result in catastrophic forgetting \cite{french1999catastrophic}, losing the knowledge it has about the older environment and task. Figure \ref{fig:figure1} shows a detailed example of such an occurrence.

To address these issues, we propose a new image coresets based selective experience replay lifelong reinforcement learning that allows DRL models to be trained on edge devices and allows DRL models to perform lifelong learning, which can learn multiple tasks without forgetting. We implemented three different image coreset implementations: neighborhood averaging coreset, neighborhood sensitivity based sampling coreset, and Maximum entropy coreset. We tested our frameworks on DIXON water and DIXON fat full body MRI images, locating five different anatomical landmark locations: left knee, right trochanter, left kidney, spleen, and lung. 

\section{Results}

\begin{figure}[htp]
    \centering
    \includegraphics[width=15cm]{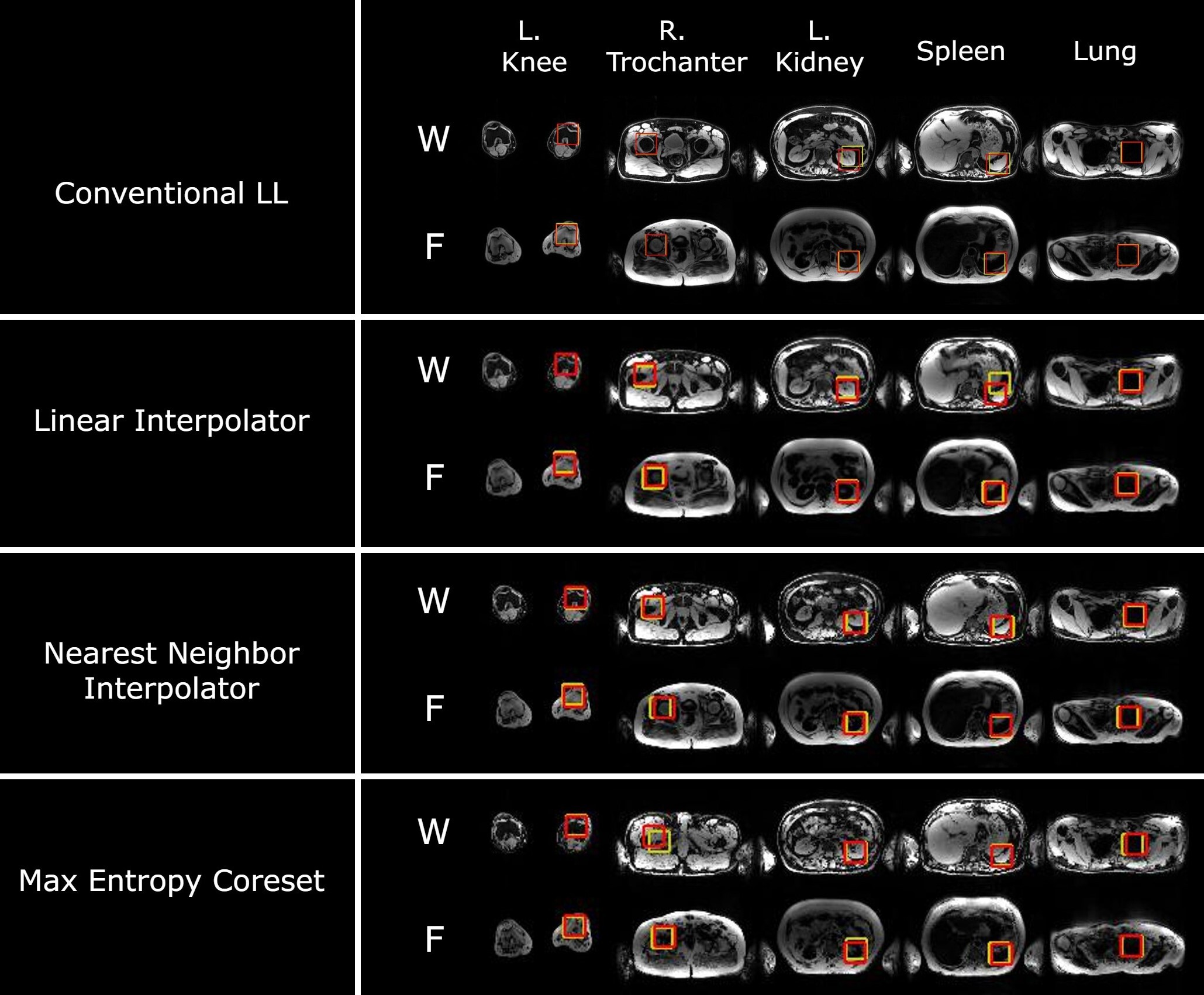}
    \caption{Illustration of the two different full-body MRI imaging environments (DIXON water denoted "W" and DIXON fat denoted "F"), and the five different anatomical landmark localization tasks (left knee, right trochanter, left kidney, spleen, and lung). The performance of the ICSERIL models with different image coreset implementations and the conventional LL model is listed. The red bounding box indicates the true landmark location and the yellow bounding box indicated the predicted landmark location of the respective models.}
    \label{fig:figure2}
\end{figure}

We trained our image coreset based selective experience replay lifelong reinforcement learning (ICSERIL) on two full-body MRI imaging environments, DIXON water, and DIXON fat, and across five different anatomical landmarks (left knee, right trochanter, left kidney, spleen, and lung). We experimented on three different image coreset implementations (neighborhood averaging, neighborhood sensitivity based sampling, and Maximum entropy). We performed 3x compression in all three dimensions of an image (x,y,z direction) using the three image coreset techniques, resulting in three different sets of 27x compression image datasets. For comparison, we trained a conventional selective experience replay lifelong reinforcement learning agent on the same dataset. The ICSERIL models with different image coreset implementations and the conventional LL model were compared for their average Euclidean distance error performance across different imaging sequences and different landmark localization tasks in the original scale without compression, examining their generalizability and learning ability on both imaging environments. Figure \ref{fig:figure2} illustrates the performance of ICSERIL models compared to a LL model across 10 different imaging environment-task pairs.

All three image coreset implementations (neighborhood averaging, neighborhood sensitivity based sampling, and Maximum entropy) demonstrated excellent performance in localizing the five different anatomical landmark locations, with $13.46\pm 14.49$, $14.17\pm 20.97$, and $11.97\pm 12.02$ respectively in the original no compression scale. They were all improved over the conventional LL model's $19.24\pm 50.77$. Maximum entropy coreset demonstrated the best average Euclidean distance error performance, the lowest standard deviation, and also made statistically significant improvement over localizing the left kidney and spleen $(p<0.05)$ in pairwise T-test compared to the conventional LL model. Table \ref{tab:table1} shows the detailed average distance error performance, pairwise T-test performance, and average pairwise T-test performance of different models.

\begin{table}[!htp]\centering
\caption{Comparison between conventional Lifelong Learning model and image coreset based selective experience replay lifelong reinforcement learning models (neighborhood averaging, neighborhood sensitivity based sampling, Maximum Entropy) for the localization of five landmarks across two imaging environments in whole-body MRI}\label{tab: }
\begin{adjustbox}{width=\textwidth}
\begin{tabular}{|l|l|l|r|r|r|r|r|}
\hline
                 &                                        & knee  & trochanter & kidney & lung  & spleen & Average \\\hline
                 & Baseline (only trained on DIXON water) & 73.44 & 45.05      & 100.33 & 40.25 & 20.63  & 55.94   \\\hline
                 & conventional LL                        & 6.34  & 30.95      & 25.50  & 21.71 & 11.69  & 19.24   \\\hline
Neighborhood Averaging           & Average distance                       & 13.66 & 8.82       & 15.37  & 18.11 & 11.32  & 13.46   \\
                 & Paired TTEST (vs. conventional LL)     & 0.13  & 0.38       & 0.04   & 0.52  & 0.92   & 0.27    \\\hline
Neighborhood Sensitivity Based Sampling & Average distance                       & 12.20 & 12.43      & 21.74  & 15.90 & 8.56   & 14.17   \\
                 & Paired TTEST (vs. conventional LL)     & 0.40  & 0.46       & 0.69   & 0.13  & 0.21   & 0.36    \\\hline
Maximum Entropy  & Average distance                       & 13.78 & 11.74      & 11.51  & 14.20 & 8.62   & 11.97   \\
                 & Paired TTEST (vs. conventional LL)     & 0.12  & 0.44       & 0.01   & 0.01  & 0.20   & 0.16  \\ \hline

\end{tabular}
\label{tab:table1}
\end{adjustbox}
\end{table}

\begin{figure}[htp]
    \centering
    \includegraphics[width=15cm]{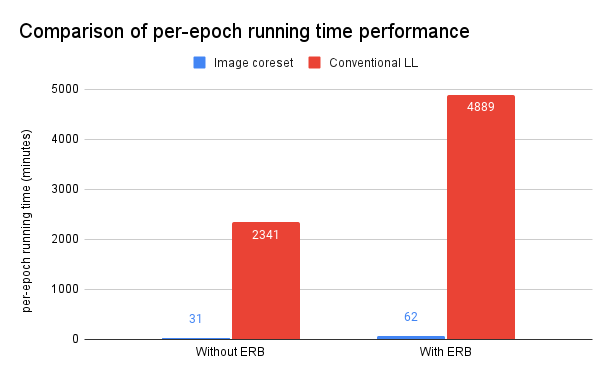}
    \caption{Illustration of per-epoch running time performance of image coreset implementations compared to the conventional LL.}
    \label{fig:figure3}
\end{figure}

Additionally, the ICSERIL models and the conventional LL model were compared for their computational complexity and computational time. With a 27x compression rate, all image coreset implementations trained significantly faster than the conventional LL model. We tested the per-epoch running time performance on a Ryzen 5 3600 with 16GB of DDR4 3200MHz RAM. All three image coreset implementations finished training an epoch without using ERBs in 31 minutes, and an epoch with using ERBs in 62 minutes. These are major speedups of 76x-79x compared to the conventional LL finishing an epoch without using ERBs in 2341 minutes, and 4889 minutes with ERBs, shown in Figure \ref{fig:figure3}. Note that the number of training epochs and rounds are the same for image coreset implementations and the conventional LL. This means that image coreset implementations were able to get better average Euclidean distance error performance shown above, while also achieving those better performances significantly faster. This shows that image coresets, especially maximum entropy coreset, can denoise the MRI images and also eliminate irrelevant information for DRL models to train more efficiently and gain better results.

\section{Discussion}
Although machine learning algorithms have massively improved the medical imaging field, researchers in the field are mostly focusing on how better and fast GPUs can speed up training or open up potential research areas over using CPU or older, less power GPUs\cite{surveyDL,KALAISELVI2017133,ParallelComputation,10.1007/978-3-642-29305-4_228,doi:10.5941/MYCO.2016.44.1.29}. However, it is crucial that medical research can be widely deployed in the real world to benefit doctors, physicians, and most importantly, patients. The medical imaging field has developed portable edge devices to collect data for a long time\cite{765266,10.1145/1579114.1579165,TROTTA2007604,doi:10.1126/sciadv.abp9307,guallart2022portable}, allowing cheaper cost and more people, geographically remote or economically disadvantaged, to benefit from significant medical breakthroughs. Additionally, if the deep learning models are deployed in the cloud using high-end GPUs, there will be serious privacy concerns, since patient data needs to be sent to the cloud, where security might be compromised and data can potentially be misused. Moreover, the latency will also be a limitation of cloud deployment. Since the data needs to be transferred to the cloud, the communication will be the bottleneck, depending on the size of the data and the communication bandwidth, the latency can be significant, while real-time feedback is essential for emergency medical applications. However, deep learning models developed for the purpose of portable edge device deployment or broad-scale deployment has been few and recent\cite{9790862,UKWANDU2022100096,s22010219}.

 Coreset has shown great results in reducing computational cost, accelerating training, or storage requirements in the medical field\cite{zheng2023selective,alexandroni2016coresets,volkov2017machine}, all of which are necessary for portable edge device deployment. 

 Lifelong learning is crucial for the edge deployment of machine learning applications to the medical field as well. With imaging environments evolving swiftly, training a model for each environment is prohibitively expensive. Not only do the models need to be stored locally on edge devices, but the user must have knowledge of which model to use. Both limit the accessibility and the hardware requirement for edge devices. Moreover, storing one model without lifelong learning for edge deployment is potentially dangerous. With catastrophic forgetting, the model will forget knowledge about previous environments, leading to incorrect prediction or evaluation, possibly causing catastrophic harm to patients.

 Our approach of incorporating image coreset techniques enables DRL models, which usually require high-end GPU or cloud computing, to be trained on a regular computer CPU. Our approach of incorporating lifelong learning enables DRL models to generalize and perform tasks in a diverse set of environments with excellent performance.

 A limitation of our framework is that we did not have much data for experiments. In the future, we plan on further testing the robustness and accuracy of our framework in more diverse imaging environments, such as pediatrics and animal images. Another limitation is that edge device models cannot share their information in the field. This is an important aspect of edge deployment since different edge devices can receive different data and information, and sharing their knowledge of new information can increase the efficiency and reliability of edge devices. Thus we need to develop methods for edge devices to share their knowledge with each other without privacy compromises.

\section{Materials and Methods}

\subsection{Lifelong deep reinforcement learning}
\begin{figure}[htp]
    \centering
    \includegraphics[width=1\linewidth]{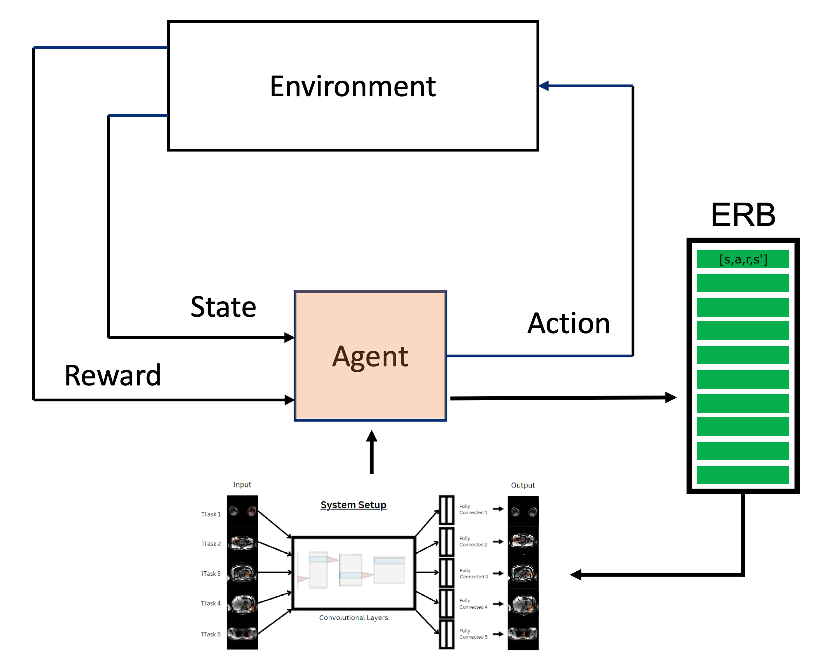}
    \caption{A schematic of the lifelong deep reinforcement learning setup for training deep reinforcement learning models. ERB stands for Experience Replay Buffer, and it contains an array of state-action-reward-resulting state tuple [s,a,r,s']}
    \label{fig:figure5}
\end{figure}

The lifelong deep reinforcement learning algorithm used in our framework was adapted based on the deep Q-Network (DQN) algorithm, shown in Figure \ref{fig:figure5}. The lifelong deep reinforcement learning algorithm is the same as Zheng et al. in their paper on ERB coresets for lifelong deep reinforcement learning \cite{Zheng23}. The environment is the 3D MRI image space, the agent is represented by a 3D bounding box that defines the state, and the potential actions it can take are going forward or backward in the x, y, or z directions. Based on the state and action, the environment outputs a reward as feedback. The reward is calculated based on the difference in distance between the agent and the target landmark bounding box before and after the action. Thus training sessions generate state-action-reward-resulting state tuples, which we sample from them to create an ERB (Experience Replay Buffer). Agents can utilize these ERBs in later training sessions to achieve lifelong learning but review previous experiences from ERBs.

\subsection{Neighborhood Averaging Coreset}
We define the neighborhood of a pixel as a $(2N-1) \times (2N-1) \times (2N-1)$ cube, where $N$ is the scaling ratio. The 3D image is split into $(X/N) \times (Y/N) \times (Z/N)$ neighborhoods, where $X,Y,Z$ are dimensions of the image. The average of all pixels in the neighborhood is taken to be in the coreset image.

\subsection{Neighborhood Sensitivity Based Sampling Coreset}
Similarly, the neighborhood definition from Neighborhood Averaging Coreset applies to Neighborhood Sensitivity Based Sampling Coreset. And the The 3D image is split into $(X/N) \times (Y/N) \times (Z/N)$ neighborhoods. Instead of computing the average of all the pixels within a neighborhood cube, we choose the central pixel of the cube as the representative for the entire neighborhood, which gets put into the coreset image. Since the dimensions of the cube are always odd numbers, denoted by $2N-1$, the center of the cube will always be a pixel.

\subsection{Maximum Entropy Coreset}
We implemented the Maximum Entropy Coreset based on the maximum entropy algorithm published in \cite{1647671}.  The formula for calculating the entropy of an image $I$ is as follows:

$$Entropy(I)=\sum_{i=0}^n p(i)log_2(p(i))$$,

where n represents the gray levels of an image, and p(i) represents the probability of a pixel in image $I$ having the gray level $i$. We assign each pixel of an image an entropy value calculated from the neighboring $10\times 10\times 10$ cube. Then we split the image into $3\times 3\times 3$ cubes and select the pixel with the highest entropy to be in the coreset image for the purpose of $3x$ scaling ratio.

\subsection{Experimental Setup}

\subsubsection{Clinical data}
We obtained DIXON water and DIXON fat images from 30 patients. We split used the first 10 patients' DIXON fat images for training, and the second 10 patients' DIXON water images for training. We combined the last 10 patients' DIXON water and DIXON fat images to produce the testing images. In summary, we have 20 training images, half is DIXON water and half is DIXON fat. We also have 20 testing images, half is DIXON water and half is DIXON fat

\subsubsection{Training protocol}
For each image coreset implementation, we trained 5 ICSERIL agents corresponding to the five landmark localization tasks (left knee, right trochanter, left kidney, spleen, and lung). The agents' states are represented as a $15\times 15 \times 9$ bounding box with. The agents trained for 4 epochs per round with a batch size of 48 for 2 rounds. The first round was trained on DIXON water and the second round was trained on DIXON fat. For comparison, we trained 5 conventional LL agents corresponding to the five landmark localization tasks as well. The agents' states are represented as a $45\times 45 \times 15$, with the rest of the training protocol identical to ICSERIL agents. We scaled the bounding box size from the original $45\times 45 \times 15$ to $15\times 15 \times 9$ because the ICSERIL models are trained on significantly smaller images. Both are trained on NVIDIA DGX-1 with four V100 GPUs.

\subsubsection{Performance Evaluation}
To compare the landmark localization performance, we used the average Euclidean distance between the model's prediction and the true target location across all environments in the original 20 test images as the metric. Since the ICSERIL models perform inference on the coreset images, their prediction $(x,y,z)$ will be multiplied by $3$ to obtain their prediction in the original images. We performed a pairwise T-test to compare the landmark localization performance of ICSERIL models to the conventional LL models. $p<0.05$ is considered a statistically significant difference.

To compare the computational efficiency performance, we used the per-epoch running time as the metric, with a lower per-epoch running time meaning signifying more computational efficiency.

\bibliographystyle{unsrtnat}
\bibliography{tz23_Image_coreset}

\begin{thebibliography}{30}
\providecommand{\natexlab}[1]{#1}
\providecommand{\url}[1]{\texttt{#1}}
\expandafter\ifx\csname urlstyle\endcsname\relax
  \providecommand{\doi}[1]{doi: #1}\else
  \providecommand{\doi}{doi: \begingroup \urlstyle{rm}\Url}\fi

\bibitem[Ghesu et~al.(2017)Ghesu, Georgescu, Zheng, Grbic, Maier, Hornegger,
  and Comaniciu]{ghesu2017multi}
Florin-Cristian Ghesu, Bogdan Georgescu, Yefeng Zheng, Sasa Grbic, Andreas
  Maier, Joachim Hornegger, and Dorin Comaniciu.
\newblock Multi-scale deep reinforcement learning for real-time 3d-landmark
  detection in ct scans.
\newblock \emph{IEEE transactions on pattern analysis and machine
  intelligence}, 41\penalty0 (1):\penalty0 176--189, 2017.

\bibitem[Tseng et~al.(2017)Tseng, Luo, Cui, Chien, Ten~Haken, and
  El~Naqa]{tseng2017deep}
Huan-Hsin Tseng, Yi~Luo, Sunan Cui, Jen-Tzung Chien, Randall~K Ten~Haken, and
  Issam El~Naqa.
\newblock Deep reinforcement learning for automated radiation adaptation in
  lung cancer.
\newblock \emph{Medical physics}, 44\penalty0 (12):\penalty0 6690--6705, 2017.

\bibitem[Maicas et~al.(2017)Maicas, Carneiro, Bradley, Nascimento, and
  Reid]{maicas2017deep}
Gabriel Maicas, Gustavo Carneiro, Andrew~P Bradley, Jacinto~C Nascimento, and
  Ian Reid.
\newblock Deep reinforcement learning for active breast lesion detection from
  dce-mri.
\newblock In \emph{International conference on medical image computing and
  computer-assisted intervention}, pages 665--673. Springer, 2017.

\bibitem[Ma et~al.(2017)Ma, Wang, Singh, Tamersoy, Chang, Wimmer, and
  Chen]{ma2017multimodal}
Kai Ma, Jiangping Wang, Vivek Singh, Birgi Tamersoy, Yao-Jen Chang, Andreas
  Wimmer, and Terrence Chen.
\newblock Multimodal image registration with deep context reinforcement
  learning.
\newblock In \emph{International Conference on Medical Image Computing and
  Computer-Assisted Intervention}, pages 240--248. Springer, 2017.

\bibitem[Alansary et~al.(2018)Alansary, Le~Folgoc, Vaillant, Oktay, Li, Bai,
  Passerat-Palmbach, Guerrero, Kamnitsas, Hou, et~al.]{alansary2018automatic}
Amir Alansary, Loic Le~Folgoc, Ghislain Vaillant, Ozan Oktay, Yuanwei Li,
  Wenjia Bai, Jonathan Passerat-Palmbach, Ricardo Guerrero, Konstantinos
  Kamnitsas, Benjamin Hou, et~al.
\newblock Automatic view planning with multi-scale deep reinforcement learning
  agents.
\newblock In \emph{International Conference on Medical Image Computing and
  Computer-Assisted Intervention}, pages 277--285. Springer, 2018.

\bibitem[Ali et~al.(2018)Ali, Hart, Gunabushanam, Liang, Muhammad, Nartowt,
  Kane, Ma, and Deng]{ali2018lung}
Issa Ali, Gregory~R Hart, Gowthaman Gunabushanam, Ying Liang, Wazir Muhammad,
  Bradley Nartowt, Michael Kane, Xiaomei Ma, and Jun Deng.
\newblock Lung nodule detection via deep reinforcement learning.
\newblock \emph{Frontiers in oncology}, 8:\penalty0 108, 2018.

\bibitem[Alansary et~al.(2019)Alansary, Oktay, Li, Le~Folgoc, Hou, Vaillant,
  Kamnitsas, Vlontzos, Glocker, Kainz, and Rueckert]{alansary2019evaluating}
Amir Alansary, Ozan Oktay, Yuanwei Li, Loic Le~Folgoc, Benjamin Hou, Ghislain
  Vaillant, Konstantinos Kamnitsas, Athanasios Vlontzos, Ben Glocker, Bernhard
  Kainz, and Daniel Rueckert.
\newblock {Evaluating Reinforcement Learning Agents for Anatomical Landmark
  Detection}.
\newblock \emph{Medical Image Analysis}, 2019.

\bibitem[Vlontzos et~al.(2019)Vlontzos, Alansary, Kamnitsas, Rueckert, and
  Kainz]{vlontzos2019multiple}
Athanasios Vlontzos, Amir Alansary, Konstantinos Kamnitsas, Daniel Rueckert,
  and Bernhard Kainz.
\newblock Multiple landmark detection using multi-agent reinforcement learning.
\newblock In \emph{International Conference on Medical Image Computing and
  Computer-Assisted Intervention}, pages 262--270. Springer, 2019.

\bibitem[Allioui et~al.(2022)Allioui, Mohammed, Benameur, Al-Khateeb,
  Abdulkareem, Garcia-Zapirain, Damaševiˇcius, and
  Maskeliunas]{Allioui2022MultAgent}
Hanane Allioui, Mazin~Abed Mohammed, Narjes Benameur, Belal Al-Khateeb,
  Karrar~Hameed Abdulkareem, Begonya Garcia-Zapirain, Robertas Damaševiˇcius,
  and Rytis Maskeliunas.
\newblock A multi-agent deep reinforcement learning approach for enhancement of
  covid-19 ct image segmentation.
\newblock \emph{Journal of Personalized Medicine}, 12:\penalty0 1--23, 2022.

\bibitem[Zhang et~al.(2021)Zhang, Du, and Liu]{Zhang2021wholeprocess}
Quan Zhang, Qian Du, and Guohua Liu.
\newblock A whole-process interpretable and multi-modal deep reinforcement
  learning for diagnosis and analysis of alzheimer's disease.
\newblock \emph{Journal of Neural Engineering}, 18:\penalty0 1--19, 2021.

\bibitem[Joseph Nathaniel~Stember(2022)]{Stember2022drl}
Hrithwik~Shalu Joseph Nathaniel~Stember.
\newblock Deep reinforcement learning with automated label extraction from
  clinical reports accurately classifies 3d mri brain volumes.
\newblock \emph{Journal of Digital Imaging}, pages 1--11, 2022.

\bibitem[French(1999)]{french1999catastrophic}
Robert~M French.
\newblock Catastrophic forgetting in connectionist networks.
\newblock \emph{Trends in cognitive sciences}, 3\penalty0 (4):\penalty0
  128--135, 1999.

\bibitem[Celard et~al.(2022)Celard, Iglesias, Sorribes-Fdez, Romero, Vieira,
  and Borrajo]{surveyDL}
Pedro Celard, Eva Iglesias, J.~Sorribes-Fdez, Rubén Romero, Seara Vieira, and
  María Borrajo.
\newblock A survey on deep learning applied to medical images: from simple
  artificial neural networks to generative models.
\newblock \emph{Neural Computing and Applications}, 35:\penalty0 1--33, 11
  2022.
\newblock \doi{10.1007/s00521-022-07953-4}.

\bibitem[Kalaiselvi et~al.(2017)Kalaiselvi, Sriramakrishnan, and
  Somasundaram]{KALAISELVI2017133}
T.~Kalaiselvi, P.~Sriramakrishnan, and K.~Somasundaram.
\newblock Survey of using gpu cuda programming model in medical image analysis.
\newblock \emph{Informatics in Medicine Unlocked}, 9:\penalty0 133--144, 2017.
\newblock ISSN 2352-9148.
\newblock \doi{https://doi.org/10.1016/j.imu.2017.08.001}.
\newblock URL
  \url{https://www.sciencedirect.com/science/article/pii/S235291481730045X}.

\bibitem[Kadah et~al.(2011)Kadah, Abd-Elmoniem, and Farag]{ParallelComputation}
Yasser Kadah, Khaled Abd-Elmoniem, and Aly Farag.
\newblock Parallel computation in medical imaging applications.
\newblock \emph{International journal of biomedical imaging}, 2011:\penalty0
  840181, 01 2011.
\newblock \doi{10.1155/2011/840181}.

\bibitem[Wang et~al.(2013)Wang, Chen, and Feng]{10.1007/978-3-642-29305-4_228}
Ning Wang, Wu-fan Chen, and Qian-jin Feng.
\newblock Angiogram images enhancement method based on gpu.
\newblock In Mian Long, editor, \emph{World Congress on Medical Physics and
  Biomedical Engineering May 26-31, 2012, Beijing, China}, pages 868--871,
  Berlin, Heidelberg, 2013. Springer Berlin Heidelberg.
\newblock ISBN 978-3-642-29305-4.

\bibitem[Nguyen et~al.(2016)Nguyen, Lee, Bae, Jeon, Mun, and
  Lee]{doi:10.5941/MYCO.2016.44.1.29}
Thi Thuong~Thuong Nguyen, Seo~Hee Lee, Sarah Bae, Sun~Jeong Jeon, Hye~Yeon Mun,
  and Hyang~Burm Lee.
\newblock Characterization of two new records of zygomycete species belonging
  to undiscovered taxa in korea.
\newblock \emph{Mycobiology}, 44\penalty0 (1):\penalty0 29--37, 2016.
\newblock \doi{10.5941/MYCO.2016.44.1.29}.
\newblock URL \url{https://doi.org/10.5941/MYCO.2016.44.1.29}.
\newblock PMID: 27103852.

\bibitem[Hwang et~al.(1998)Hwang, Quistgaard, Souquet, and Crum]{765266}
Juin-Jet Hwang, J.~Quistgaard, J.~Souquet, and L.A. Crum.
\newblock Portable ultrasound device for battlefield trauma.
\newblock In \emph{1998 IEEE Ultrasonics Symposium. Proceedings (Cat. No.
  98CH36102)}, volume~2, pages 1663--1667 vol.2, 1998.
\newblock \doi{10.1109/ULTSYM.1998.765266}.

\bibitem[Sachpazidis et~al.(2009)Sachpazidis, Kontaxakis, and
  Sakas]{10.1145/1579114.1579165}
Ilias Sachpazidis, George Kontaxakis, and Georgios Sakas.
\newblock A portable medical unit for medical imaging telecollaboration.
\newblock In \emph{Proceedings of the 2nd International Conference on PErvasive
  Technologies Related to Assistive Environments}, PETRA '09, New York, NY,
  USA, 2009. Association for Computing Machinery.
\newblock ISBN 9781605584096.
\newblock \doi{10.1145/1579114.1579165}.
\newblock URL \url{https://doi.org/10.1145/1579114.1579165}.

\bibitem[Trotta et~al.(2007)Trotta, Massari, Palermo, Scopinaro, and
  Soluri]{TROTTA2007604}
C.~Trotta, R.~Massari, N.~Palermo, F.~Scopinaro, and A.~Soluri.
\newblock New high spatial resolution portable camera in medical imaging.
\newblock \emph{Nuclear Instruments and Methods in Physics Research Section A:
  Accelerators, Spectrometers, Detectors and Associated Equipment},
  577\penalty0 (3):\penalty0 604--610, 2007.
\newblock ISSN 0168-9002.
\newblock \doi{https://doi.org/10.1016/j.nima.2007.03.037}.
\newblock URL
  \url{https://www.sciencedirect.com/science/article/pii/S0168900207005268}.

\bibitem[Basser(2022)]{doi:10.1126/sciadv.abp9307}
Peter Basser.
\newblock Detection of stroke by portable, low-field mri: A milestone in
  medical imaging.
\newblock \emph{Science Advances}, 8\penalty0 (16):\penalty0 eabp9307, 2022.
\newblock \doi{10.1126/sciadv.abp9307}.
\newblock URL \url{https://www.science.org/doi/abs/10.1126/sciadv.abp9307}.

\bibitem[Guallart-Naval et~al.(2022)Guallart-Naval, Algar{\'\i}n,
  Pellicer-Guridi, Galve, Vives-Gilabert, Bosch, Pall{\'a}s, Gonz{\'a}lez,
  Rigla, Mart{\'\i}nez, et~al.]{guallart2022portable}
Teresa Guallart-Naval, Jos{\'e}~M Algar{\'\i}n, Rub{\'e}n Pellicer-Guridi,
  Fernando Galve, Yolanda Vives-Gilabert, Rub{\'e}n Bosch, Eduardo Pall{\'a}s,
  Jos{\'e}~M Gonz{\'a}lez, Juan~P Rigla, Pablo Mart{\'\i}nez, et~al.
\newblock Portable magnetic resonance imaging of patients indoors, outdoors and
  at home.
\newblock \emph{Scientific Reports}, 12\penalty0 (1):\penalty0 13147, 2022.

\bibitem[Ali et~al.(2022)Ali, Ali, Shah, and Shahzad]{9790862}
Owais Ali, Hazrat Ali, Syed Ayaz~Ali Shah, and Aamir Shahzad.
\newblock Implementation of a modified u-net for medical image segmentation on
  edge devices.
\newblock \emph{IEEE Transactions on Circuits and Systems II: Express Briefs},
  69\penalty0 (11):\penalty0 4593--4597, 2022.
\newblock \doi{10.1109/TCSII.2022.3181132}.

\bibitem[Ukwandu et~al.(2022)Ukwandu, Hindy, and Ukwandu]{UKWANDU2022100096}
Ogechukwu Ukwandu, Hanan Hindy, and Elochukwu Ukwandu.
\newblock An evaluation of lightweight deep learning techniques in medical
  imaging for high precision covid-19 diagnostics.
\newblock \emph{Healthcare Analytics}, 2:\penalty0 100096, 2022.
\newblock ISSN 2772-4425.
\newblock \doi{https://doi.org/10.1016/j.health.2022.100096}.
\newblock URL
  \url{https://www.sciencedirect.com/science/article/pii/S2772442522000417}.

\bibitem[Garifulla et~al.(2022)Garifulla, Shin, Kim, Kim, Kim, Kim, and
  Hong]{s22010219}
Mukhammed Garifulla, Juncheol Shin, Chanho Kim, Won~Hwa Kim, Hye~Jung Kim,
  Jaeil Kim, and Seokin Hong.
\newblock A case study of quantizing convolutional neural networks for fast
  disease diagnosis on portable medical devices.
\newblock \emph{Sensors}, 22\penalty0 (1), 2022.
\newblock ISSN 1424-8220.
\newblock \doi{10.3390/s22010219}.
\newblock URL \url{https://www.mdpi.com/1424-8220/22/1/219}.

\bibitem[Zheng et~al.(2023{\natexlab{a}})Zheng, Zhou, Braverman, Jacobs, and
  Parekh]{zheng2023selective}
Guangyao Zheng, Samson Zhou, Vladimir Braverman, Michael~A Jacobs, and Vishwa~S
  Parekh.
\newblock Selective experience replay compression using coresets for lifelong
  deep reinforcement learning in medical imaging.
\newblock \emph{arXiv preprint arXiv:2302.11510}, 2023{\natexlab{a}}.

\bibitem[Alexandroni et~al.(2016)Alexandroni, Moreno, Sochen, and
  Greenspan]{alexandroni2016coresets}
Guy Alexandroni, Gali~Zimmerman Moreno, Nir Sochen, and Hayit Greenspan.
\newblock Coresets vs clustering: comparison of methods for redundancy
  reduction in very large white matter fiber sets.
\newblock In \emph{Medical Imaging 2016: Image Processing}, volume 9784, pages
  65--73. SPIE, 2016.

\bibitem[Volkov et~al.(2017)Volkov, Hashimoto, Rosman, Meireles, and
  Rus]{volkov2017machine}
Mikhail Volkov, Daniel~A Hashimoto, Guy Rosman, Ozanan~R Meireles, and Daniela
  Rus.
\newblock Machine learning and coresets for automated real-time video
  segmentation of laparoscopic and robot-assisted surgery.
\newblock In \emph{2017 IEEE international conference on robotics and
  automation (ICRA)}, pages 754--759. IEEE, 2017.

\bibitem[Zheng et~al.(2023{\natexlab{b}})Zheng, Zhou, Braverman, Jacobs, and
  Parekh]{Zheng23}
Guangyao Zheng, Samson Zhou, Vladimir Braverman, Michael~A. Jacobs, and
  Vishwa~S. Parekh.
\newblock Selective experience replay compression using coresets for lifelong
  deep reinforcement learning in medical imaging, 2023{\natexlab{b}}.

\bibitem[Kao and Nutter(2006)]{1647671}
P.B. Kao and B.~Nutter.
\newblock Application of maximum entropy-based image resizing to biomedical
  imaging.
\newblock In \emph{19th IEEE Symposium on Computer-Based Medical Systems
  (CBMS'06)}, pages 813--819, 2006.
\newblock \doi{10.1109/CBMS.2006.46}.

\end{thebibliography}
\end{document}